\documentclass[fleqn,10pt]{wlscirep}
\usepackage[utf8]{inputenc}
\usepackage[T1]{fontenc}
\usepackage[linesnumbered,ruled,commentsnumbered]{algorithm2e}
\usepackage{hyperref}
\usepackage{ragged2e}

\title{Complementary Learning System Empowers Online Continual Learning of Vehicle Motion Forecasting in Smart Cities}

\author[1]{Zirui Li}
\author[1]{Yunlong Lin}
\author[1]{Guodong Du}
\author[2]{Xiaocong Zhao}
\author[1]{Cheng Gong}
\author[3]{Chen Lv}
\author[1,*]{Chao Lu}
\author[1,*]{Jianwei Gong}
\affil[1]{The School of Mechanical Engineering, Beijing Institute of Technology, Beijing 100081, China.}
\affil[2]{The Key Laboratory of Road and Traffic Engineering, Ministry of Education, Tongji University, Shanghai, China.}
\affil[3]{The School of Mechanical and Aerospace Engineering, Nanyang Technological University, Singapore.}
\affil[*]{Corresponding authors: chaolu@bit.edu.cn (C. Lu) and gongjianwei@bit.edu.cn (J. Gong)}


\begin{abstract}

Artificial intelligence underpins most smart city services, yet deep neural network (DNN) that forecasts vehicle motion still struggle with catastrophic forgetting, the loss of earlier knowledge when models are updated. Conventional fixes enlarge the training set or replay past data, but these strategies incur high data collection costs, sample inefficiently and fail to balance long- and short-term experience, leaving them short of human-like continual learning. Here we introduce Dual-LS, a task-free, online continual learning paradigm for DNN-based motion forecasting that is inspired by the complementary learning system of the human brain. Dual-LS pairs two synergistic memory rehearsal replay mechanisms to accelerate experience retrieval while dynamically coordinating long-term and short-term knowledge representations. Tests on naturalistic  data spanning three countries, over 772,000 vehicles and cumulative testing mileage of 11,187 km show that Dual-LS mitigates catastrophic forgetting  by up to 74.31\% and reduces computational resource demand by up to 94.02\%, markedly boosting predictive stability in vehicle motion forecasting without inflating data requirements. Meanwhile, it endows DNN-based vehicle motion forecasting with computation efficient and human-like continual learning adaptability fit for smart cities.

\end{abstract}
\begin{document}

\flushbottom
\maketitle
% \footnotetext{Under formal peer review at Nature Computational Science}

% Nature Machine Intelligence

\thispagestyle{empty}


\section*{Introduction}

With advances in deep learning and high-performance computing, artificial intelligence (AI) has achieved remarkable breakthroughs\cite{mnih2015human,silver2018general,jumper2021highly}. To boost operational efficiency and reduce human workload, many data-driven algorithms have been embedded in smart cities\cite{feng2023dense,kaufmann2023champion,song2023reaching,gehrig2024low}. Vehicle motion forecasting\cite{mozaffari2020deep}, as a typical technology of intelligent vehicles application in smart cities, plays an important role in alleviating traffic congestion\cite{wang2024traffic}, protecting the safety of vulnerable road users\cite{jafari2024pedestrians}, improving driving comfort\cite{chen2025predicting} and reducing accident rates\cite{cao2023continuous}. Because naturalistic scenarios evolve rapidly and future conditions are difficult to predict, these systems must learn and adapt continuously throughout their lifecycles\cite{lesort2020continual}. Nowadays, richer collected data  and stronger processing capabilities allow most AI modules to be trained on ever-larger datasets to improve performance. This trend, however, raises two challenges: storing and maintaining vast sensor archives is costly, and collecting all possible cases in a single effort is unrealistic. A common workaround is to train an initial model and then update it incrementally as new data arrive\cite{tajbakhsh2016convolutional}. From the perspective of DNN, such continual updates can trigger catastrophic forgetting, the gradual loss of proficiency on previously learned tasks\cite{perkonigg2021dynamic}.

To maintain the memory stability of DNN and mitigate catastrophic forgetting under real-world data streams, researchers have advanced continual learning (CL) techniques\cite{van2022three,kirkpatrick2017overcoming}. CL seeks to endow models with the dual capacity to assimilate new tasks while retaining competence on previously learned ones during continuous updates. Within the CL framework, assorted paradigms and algorithms have emerged, each tailored to specific assumptions, such as whether task labels are supplied, how much memory can be stored, the desired learning efficiency, and the number of classes involved \cite{wang2024comprehensive}. These assumptions give rise to replay-based, generative-based and dynamic architecture strategies\cite{wang2024comprehensive}. Most current CL methods, however, are validated only on simplified benchmarks\cite{de2021continual}, restricting their usefulness in safety-critical transportation systems\cite{lesort2020continual}. In self-driving vehicles (SVs), DNN-based motion forecasting, which predicts the future trajectories of surrounding agents and feeds this information directly into the planning stack, is a linchpin of overall safety and performance. Enhancing the CL capacity of this component can preserve prediction fidelity in the face of non-stationary traffic patterns, thereby protecting vulnerable road users, improving traffic flow and reducing conflict incidents\cite{cao2023continuous}.

\begin{figure}[t!]
\centering
\includegraphics[width=0.98\linewidth]{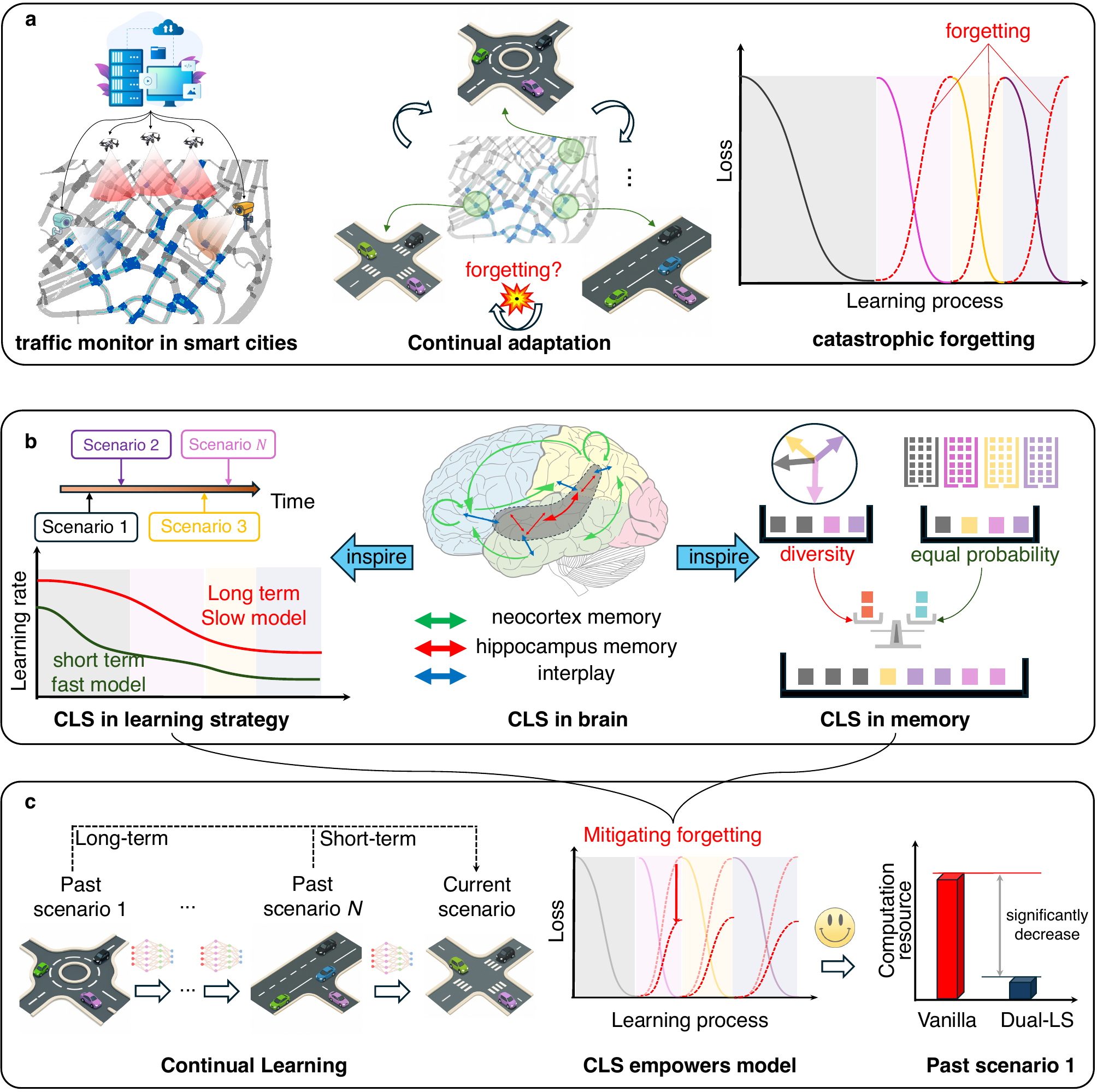}
\caption{\justifying The proposed CLS-inspired Dual-LS algorithm. \textbf{a}, The DNN model is sequentially trained by the samples from each scenario and trying to adapt to the newly encounter one. It may suffer from the catastrophic forgetting when meets the previously learnt scenario. Left: the traffic monitor system in smart cities; Middle: the DNN-based system continual adaptation to various traffic scenarios; Right: the catastrophic forgeting occurs in the learning process. \textbf{b}, The CLS in brain learning strategy and replay memory buffer. 
Left: The long-short term learning strategy with long-term memory slow model and short-term memory fast model. The cooperation of these two models mimics the interplay mechanism between hippocampus and neocortex. Middle: the CLS between neocortex and hippocampus in the human brain. Right: the dual memory buffer with two kinds of samples from reservoir sampling and gradient-based diversity sampling is designed, which serves for a better trade-off between equal probability-based and diversity enhanced-based sampling strategies. \textbf{c}, The synthesis of dual memory and the long-short term learning strategy empowers online continual learning. The catastrophic forgetting is mitigated and the demand of computational resource is reduced.}
\label{fig1_framework}
\end{figure}

However, there are two major shortcomings in current attempts. First, there is a high reliance on task identifiers to indicate boundaries between tasks, which are hard to define accurately or meaningfully; second, memory samples from the learned tasks are typically selected randomly, lacking a refined maintenance and efficient utilization strategy. In this article, to address the first issue, we propose a modified paradigm for DNN-based modules with online task-free continual learning (OTFCL) ability. This paradigm consists of four key components: data stream, memory storage, model update strategies, and performance evaluation. Similar to the learning process of the human brain, OTFCL is required to integrate new experiences with prior knowledge to update its cognitive abilities and form memories. When the model meets cases it has learned, the performance can reflect whether catastrophic forgetting happens. Specifically, the characteristics of each component in the OTFCL paradigm are as follows:
\begin{itemize}
    \item \textbf{Data stream}: The data stream in CL is a dynamic, evolving, and potentially non-repetitive flow of information. The DNN-based models need to adapt to the new distributions in the data stream while retaining old knowledge from past information.
    \item \textbf{Memory storage}: The memory storage refers to the mechanisms used to retain and recall knowledge from previous tasks or experiences to mitigate catastrophic forgetting. It plays a critical role in CL as models are typically limited in terms of how much past data they can store or replay.
    \item \textbf{Model update strategies}: The strategy is the algorithm for weight update in DNN by complying with the constraints in the memory limit, the learning setting, and the related assumption. The goal of the strategy is to improve the performance expectations within OTFCL. 
    \item \textbf{Performance evaluation}: Besides the final accuracy after the whole learning process, the evaluation metrics need to measure the level of forgetting, which is also termed memory stability.
\end{itemize}

In the OTFCL paradigm, any method that can leverage the data stream to update models can be incorporated, such as continually training without recalling past samples and updating the model with human feedback. In the fields of computer vision and reinforcement learning, several algorithms have been proposed\cite{wang2024comprehensive}, including gradient projection, rehearsal replay, generative replay, and the integration of these approaches. Experimentally, we found that the rehearsal replay-based methods performed better than others\cite{buzzega2020dark}. However, we also identified that during memory replay, the reliance on random sampling with equal probability lacks effective memory management, which is the second shortcoming. It can result in important samples, that may be critical for mitigating forgetting, not being selected during the historical sample retrieval process. Additionally, during the process of model updates, all samples are treated equally, without distinguishing between short-term quick memories and long-term foundational memories in the algorithmic design. Inspired by the complex neural connections in the human brain\cite{kudithipudi2022biological}, numerous DNN architectures have been designed to tackle various tasks\cite{wang2024comprehensive}. For CL, the recent trend in neuroscience has demonstrated that the human brain's memory maintenance and knowledge accumulation align with the Complementary Learning System (CLS) theory\cite{kumaran2016learning,sun2023organizing}. According to the CLS theory\cite{sun2023organizing}, effective learning requires two complementary systems: one, located in the neocortex, serves as the basis for the gradual acquisition of structured knowledge about the environment, while the other, centered on the hippocampus, allows rapid learning of the specifics of individual items and experiences. Inspired by the biological mechanism of CLS theory, we propose a novel algorithm with \textbf{Dual} memory buffer and \textbf{L}ong-\textbf{S}hort term learning strategy under the OTFCL paradigm, which is termed Dual-LS.

\begin{itemize}
    \item \textbf{Dual memory buffer}: In the replay-based CL, a new memory buffer with two kinds of samples from reservoir sampling and gradient-based diversity sampling\cite{aljundi2019gradient}, which serve for equal probability in the data stream and are immune to the imbalanced data stream, respectively. 
    \item \textbf{Long-short term learning strategy}: Inspired by the CLS theory in the human brain\cite{kumaran2016learning}, a new model update strategy is developed to trade-off the long-term consolidate and short-term rapid knowledge, which mimics the interplay mechanism between hippocampus and neocortex.
\end{itemize}

To demonstrate the effectiveness of the designed paradigm and the proposed algorithm for CL, the vehicle motion forescasting\cite{mozaffari2020deep} is selected as the learning objective, which is a typical, challenging and safe-related task for SV. It is to predict the multi-modal behaviours of the target vehicle in the environment with multiple interactive vehicles. Accurate prediction of the target vehicle's future behavior can support the SV’s reasonable decision-making and safe navigation. For the OTFCL paradigm, the data stream is created based on the worldwide naturalistic driving data with over 772,000 vehicles and cumulative testing mileage of 11,187 km. The dataset consists of multiple interactive traffic scenarios, such as the highway on-ramp, intersections, roundabouts, etc. Several metrics are developed to quantify CL performance in terms of adaptability to new tasks and the ability to mitigate catastrophic forgetting. Comparative experiments with multiple baseline methods demonstrated that the proposed method exhibits excellent performance in vehicle motion forescasting. From a CL perspective, Dual-LS can maintain memory stability and learning efficiency, simultaneously. To the best of our knowledge, this is the first systematic CL paradigm for vehicle motion forecasting in smart cities. Additionally, the proposed Dual-LS provides an outlook for biological intelligence-inspired CL algorithms. In this article, the paradigm, algorithm, and validation pipeline are versatile, and not only applicable to DNN-based modules in SVs but also to other data-driven  systems\cite{yan2023learning,feng2021intelligent,yu2024online}. For example, they could be applied to the self-evolving system\cite{wang2024traffic,duan2023spatiotemporal} or to evaluate the CL capability of smart cities\cite{doshi2020continual}.

\section*{Results}

\begin{figure}[t!]
\centering
\includegraphics[width=0.85\linewidth]{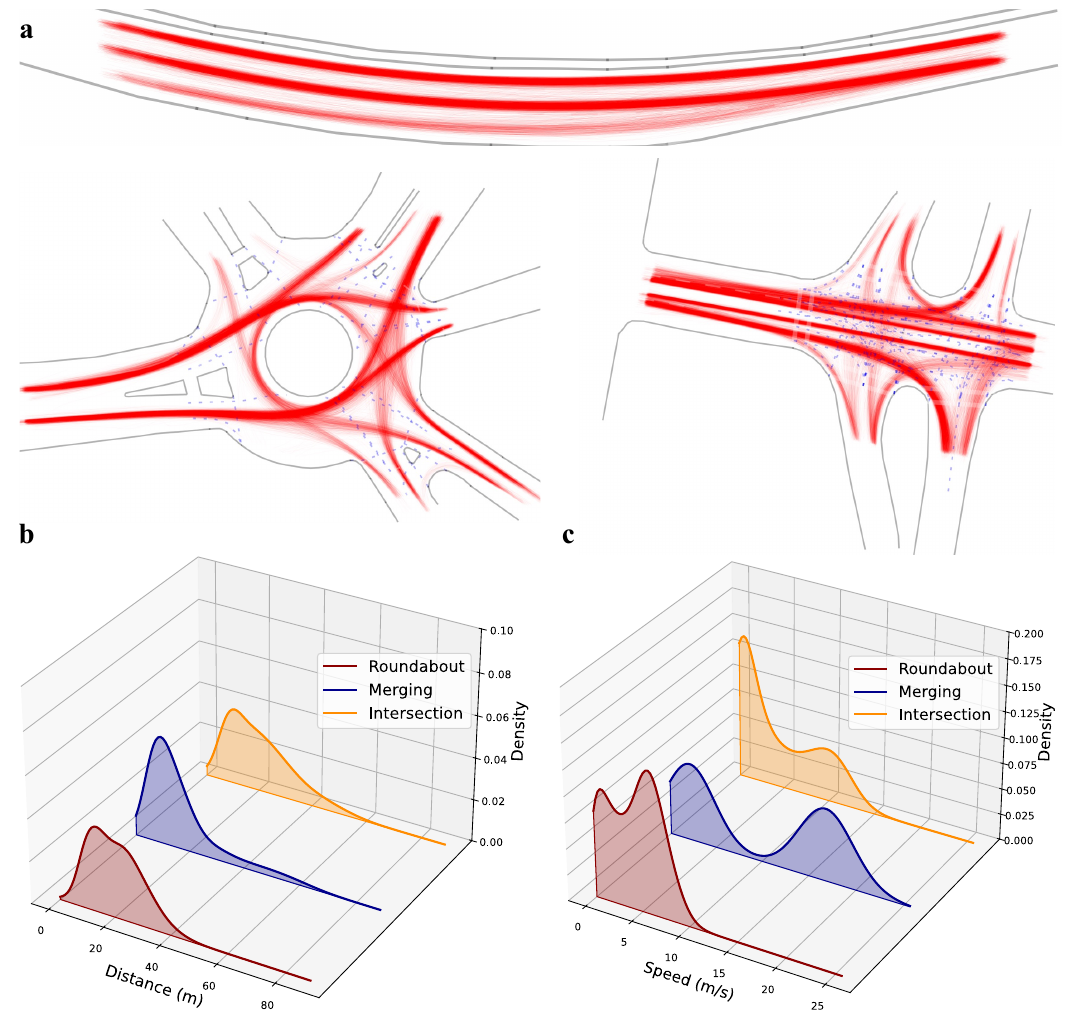}
\caption{\justifying The illustration of the used dataset\cite{zhan2019interaction}. Three scenarios (roundabout, merging and intersection) are selected for comparison. \textbf{a}, The visualization of map and all trajectories in the training set. \textbf{b}, The distributions of distance between the target vehicle and surrounding vehicles. \textbf{c}, The distributions of distance between the vehicle speed. The difference in trajectories, relative distance and speed indicates that these three scenarios are distinguished with each other.}
\label{fig2_dataset}
\end{figure}

\subsection*{Dataset description}
In this paper, the DNN of smart cities is specified as the motion forecasting of SVs. To effectively validate the performance of continual learning, the INTERACTION dataset is employed\cite{zhan2019interaction}, which clearly categorizes data collected by unmanned aerial vehicles across multiple distinct scenarios. It covers several countries and includes diverse scenarios such as intersections, roundabouts, and highways. Compared to datasets that do not differentiate between scenarios (e.g., Waymo\cite{ettinger2021large}), this is particularly beneficial for designing experiments related to DIL\cite{van2022three}. Although the proposed Dual-LS is task-free, meaning that it does not require the information of task boundary, the clear segmentation of tasks aids in standardizing tests and ensuring fair comparisons. According to the order in which the data of different scenarios appear in the data stream, they are outlined as: intersection\#1 (MA), roundabout\#1 (FT), roundabout\#2 (LN), Merging\#1 (ZS2), roundabout\#3 (OF), intersection\#2 (EP0), intersection\#3 (GL), Merging\#2 (ZS0). The number of samples in each scenarios are provided in the Supplementary Material (Section 1). Fig.~\ref{fig2_dataset}a illustrates the differences in feature distribution across different scenarios, using scenes ZS2, FT, and GL as examples. As shown in Fig.~\ref{fig2_dataset}b and Fig.~\ref{fig2_dataset}c, the distributions of relative distance and vehicle speed vary significantly across different scenarios. The statistical results of remaining scenarios are listed in the Supplementary Material (Section 1).

\subsection*{Mitigating memory and computational resource demand}

\begin{figure}[h!]
\centering
\includegraphics[width=0.98\linewidth]{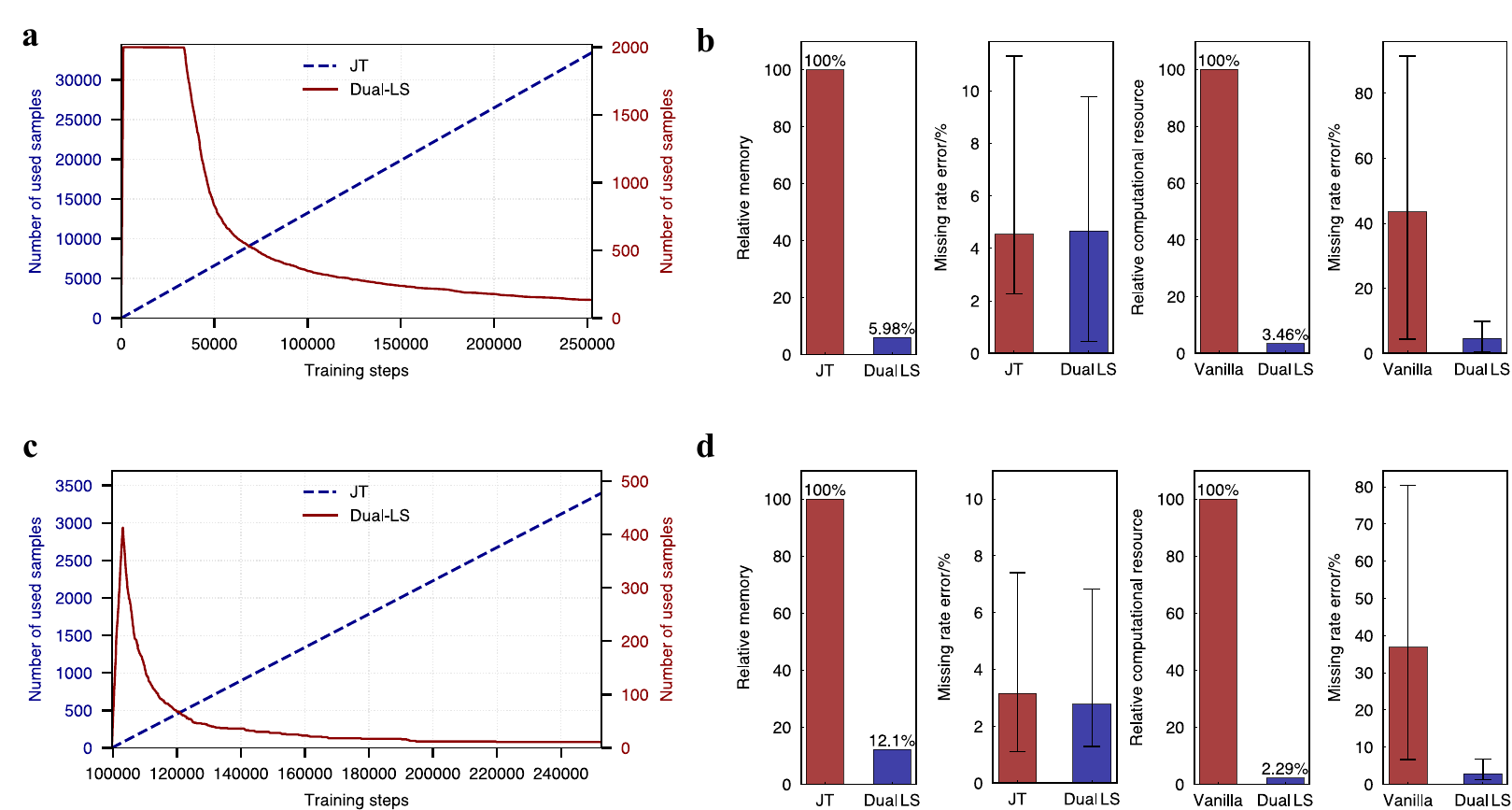}
\caption{\justifying The overall performance of Dual-LS, Vanilla and joint training in mitigating catastrophic forgetting, memory and computational resource demand. \textbf{a} and \textbf{c}, the cumulative volume of data processed over the entire training procedure for Task 1 Task 3, respectively. \textbf{b} and \textbf{d} (from left to right), the relative memory used between JT and Dual-LS, the missing rate error (\%) between JT and Dual-LS, the relative computational resource used between Vanilla and Dual-LS, the missing rate error (\%) between Vanilla and Dual-LS. The error bars represent the standard error of 10 runs.}
\label{fig3_cost}
\end{figure}

Continual learning algorithms are valuable because they maintain motion forecasting performance while restraining memory and compute demands. Fig.~\ref{fig3_cost} depicts prediction error alongside relative consumption resource. The joint training (JT) is trained on all data at once, whereas Vanilla proceeds training sequentially from task 1 to 8. For task 1 (Fig.~\ref{fig3_cost}a and b), Dual-LS cuts the peak memory requirement by 94.02\% compared with JT, without materially affecting prediction error. Fig.~\ref{fig3_cost}a plots the cumulative volume of data processed over the entire training procedure, a quantity that is broadly proportional to the peak memory footprint required. Compared to Vanilla, Dual-LS lowers both prediction error and computational cost by 96.54\% and 34.08\%, respectively, demonstrating that it guards past task 1 against catastrophic forgetting even under a markedly slimmer computational budget. Similar trends hold for task 3 (Fig.~\ref{fig3_cost}c and d). Collectively, these results show that Dual-LS simultaneously mitigates catastrophic forgetting and curbs memory and compute overhead.

\subsection*{Mitigating catastrophic forgetting}

\begin{figure}[h!]
    \centering
    \includegraphics[width=1\linewidth]{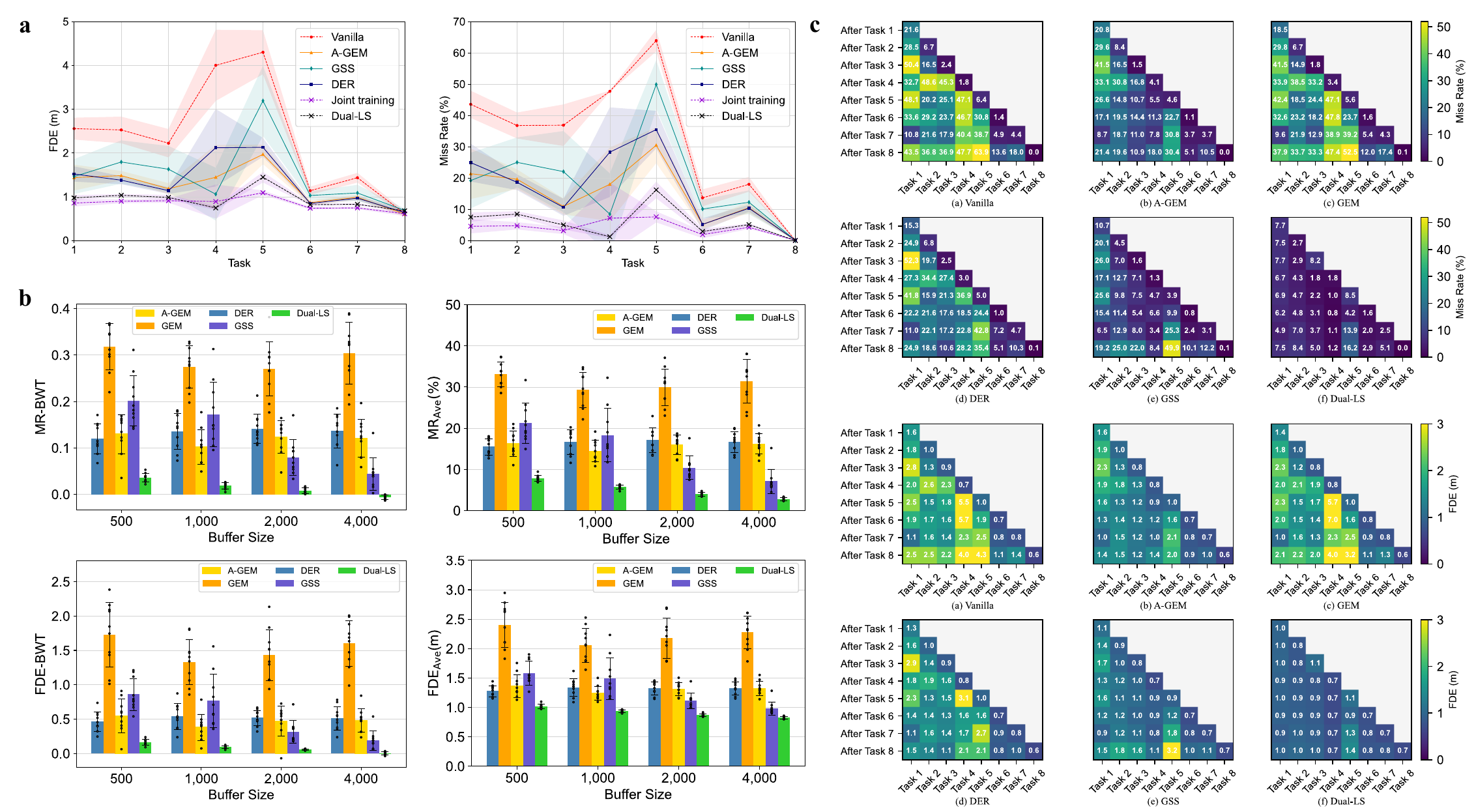}
\caption{\justifying The overall results of Dual-LS and other baseline methods in mitigating the catastrophic forgetting. \textbf{a}, The quantification of catastrophic forgetting in FDE and MR (Total memory buffer size: 1000). The x-axis represents that each model is trained from Task 1 to Task 8. The y-axis is the metrics. The transparent area is the standard deviation of 10 repeated experiments. \textbf{b}, The comparison of all methods in different total memory buffer size (500, 1000, 2000 and 4000) with several metrics (MR-BWT, FDE-BWT, MR-AVE, FDE- AVE). The error bars donate the standard deviation of 10 repeated experiments, while scatters represent each experiment’s results. \textbf{c}, The matrix of errors (FDE and MR) in the process of CL (Total memory buffer size: 1000). The $(i,j)\textsuperscript{th}$ in the matrix represents the error when model is trained after Task $i$ and then tested in Task $j$. The error matrix reflects the detailed variation in memory stability.}
    \label{fig4_statistic_results}
\end{figure}

Fig.~\ref{fig4_statistic_results} presents the overall results of the Dual-LS algorithm compared to all baseline methods. It is important to note that, while Fig.~\ref{fig4_statistic_results} explicitly delineates different tasks, the task index and boundary information were not disclosed to algorithms. This division is solely for facilitating standardized comparative studies. As shown in Fig.~\ref{fig4_statistic_results}b, for the four metrics (MR, FDE, MR-BWT, and FDE-BWT), CL-based algorithms outperform non-CL algorithm (Vanilla). This confirms the existence of catastrophic forgetting and demonstrates that CL algorithms can mitigate it effectively. Joint training represents the theoretical error lower bound, and the experimental results indicate that all CL methods achieve performance closer to joint training than Vanilla. After learning all 8 tasks, both non-CL and CL algorithms exhibit higher prediction errors on Task 4 and Task 5. This may be due to the significant differences between these two tasks and the others. Among all CL methods, the Dual-LS algorithm achieves lower prediction errors compared to the other three CL algorithms (A-GEM\cite{chaudhry2018efficient}, GSS\cite{aljundi2019gradient}, and DER\cite{buzzega2020dark}), with performance closer to joint training, the theoretical lower bound. This highlights the potential of the CLS theory in biological intelligence to enhance the predictive performance of CL models. In Fig.~\ref{fig4_statistic_results}b, after all algorithms complete Task 8, the performance on this recently learned task is comparable, with low errors across the board. This indicates that all algorithms adapt well to the current task. However, CL algorithms uniquely maintain adaptability to the current task while alleviating catastrophic forgetting of past tasks.

Similar to the memory replay mechanism in human learning, buffer memory replay serves as the foundation of the proposed OTFCL paradigm, and its effectiveness is influenced by the buffer size. Fig.~\ref{fig4_statistic_results}a illustrates the average performance of all CL algorithms under four different buffer sizes. As shown in the Fig.~\ref{fig4_statistic_results}b, Dual-LS outperforms all other CL methods and achieves optimal results across different experimental settings. Additionally, the results for GSS and Dual-LS demonstrate that increasing the buffer size enhances model performance. This phenomenon aligns with observations in human learning and memory processes: having a larger memory capacity reduces the forgetting of previously acquired knowledge. This indicates that, compared to A-GEM and DER, both GSS and Dual-LS can effectively utilize the memory buffer. Each bar in Fig.~\ref{fig4_statistic_results}b not only presents the average results but also includes the standard deviation from 10 repeated experiments. Fig.~\ref{fig4_statistic_results}a shows that the Dual-LS algorithm exhibits more stable predictive performance, achieving both lower average error and smaller standard deviation.

To better capture the dynamics of CL algorithms across sequential tasks, Fig.~\ref{fig4_statistic_results}c illustrates the performance of models on all past tasks' test sets after completing each new task. Overall, it can be observed that with the completion of each new task, the performance on past tasks deteriorates for all algorithms, which reflects the phenomenon of catastrophic forgetting. However, compared to the vanilla algorithm, all five CL algorithms mitigate this issue to some extent. From the stepwise plots in Fig.~\ref{fig4_statistic_results}c, it is evident that after learning each new task, Dual-LS consistently achieves better and more stable performance on past tasks compared to A-GEM, GEM, GSS, and DER in most cases. More results of comparative results are shown in Supplementary Material (Section 5).

\begin{figure}[htbp!]
\centering
\includegraphics[width=0.90\linewidth]{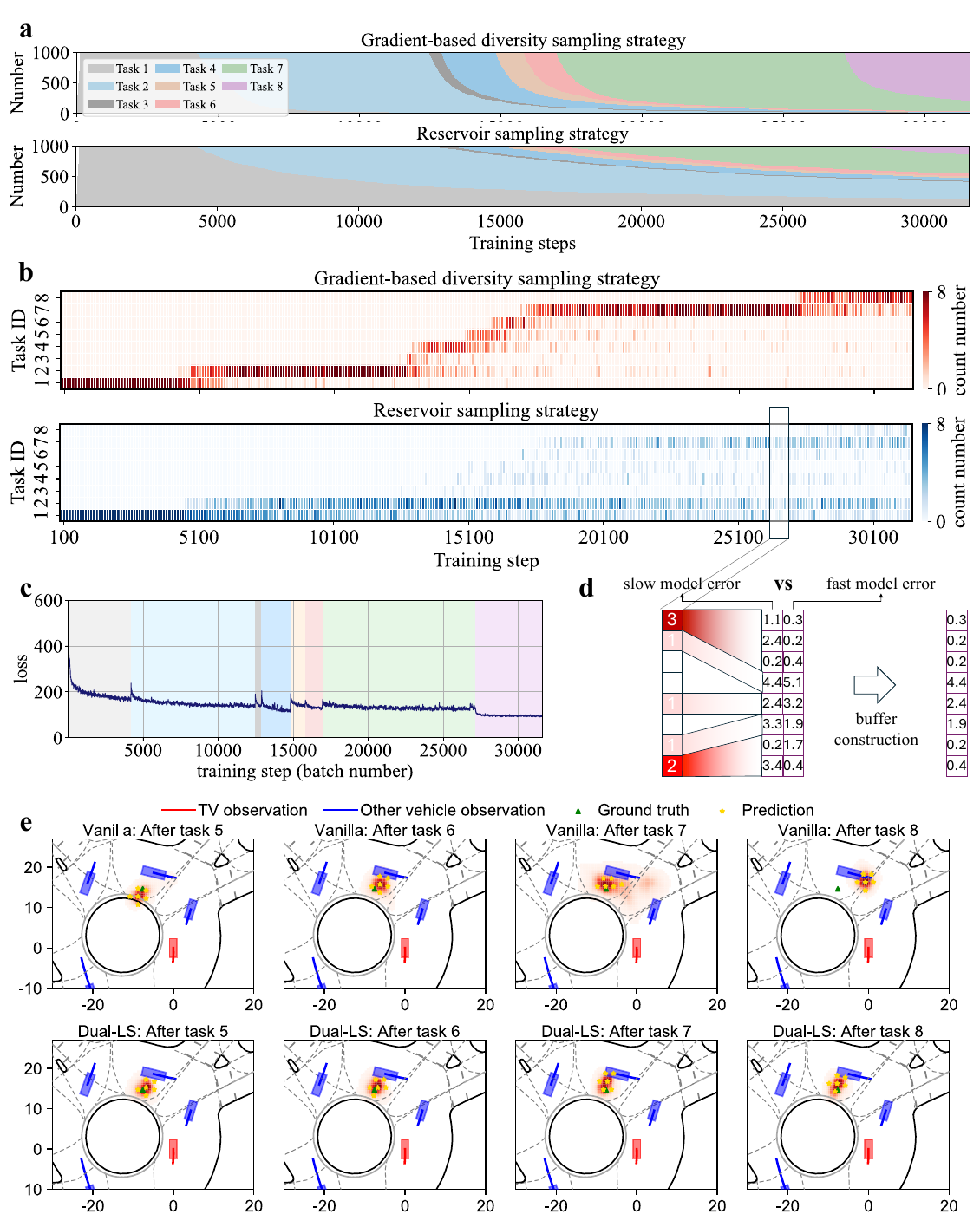}
\caption{\justifying The CLS-inspired learning mechanism of Dual-LS in the CL setting. \textbf{a}, The proportion of samples from each past tasks in the memory buffer. It changes after each update of memory buffer. The comparison of two kinds of sampling strategies highlights the significant of CLS in memory buffer design. \textbf{b}, The constitution of two memory buffers from each past task. The color represents the number of samples from corresponding past task. \textbf{c}, The variation of training loss. Each background color corresponds to a task. \textbf{d}, The illustration of long-short term learning strategy. Each sample in the memory buffer has two errors from slow and fast models, respectively.  \textbf{e}, Case study in the test set from scenario FT (task 2). Top: the vanilla model trained after 5-8 tasks. Down: the Dual-LS trained after 5-8 tasks. In each case, the predicions are represented by six yellow stars in red heatmaps, where the specific predicted postions are sampled based on the predicted heatmap.}
\label{fig5_CLS}
\end{figure}

\subsection*{The effect of complementary learning system}
The above analysis primarily focuses on the overall performance improvements brought by the CLS-inspired algorithm to CL. This section delves into the mechanisms behind how Dual-LS embodies principles of biological intelligence.

\begin{figure}[t!]
    \centering
    \includegraphics[width=1\linewidth]{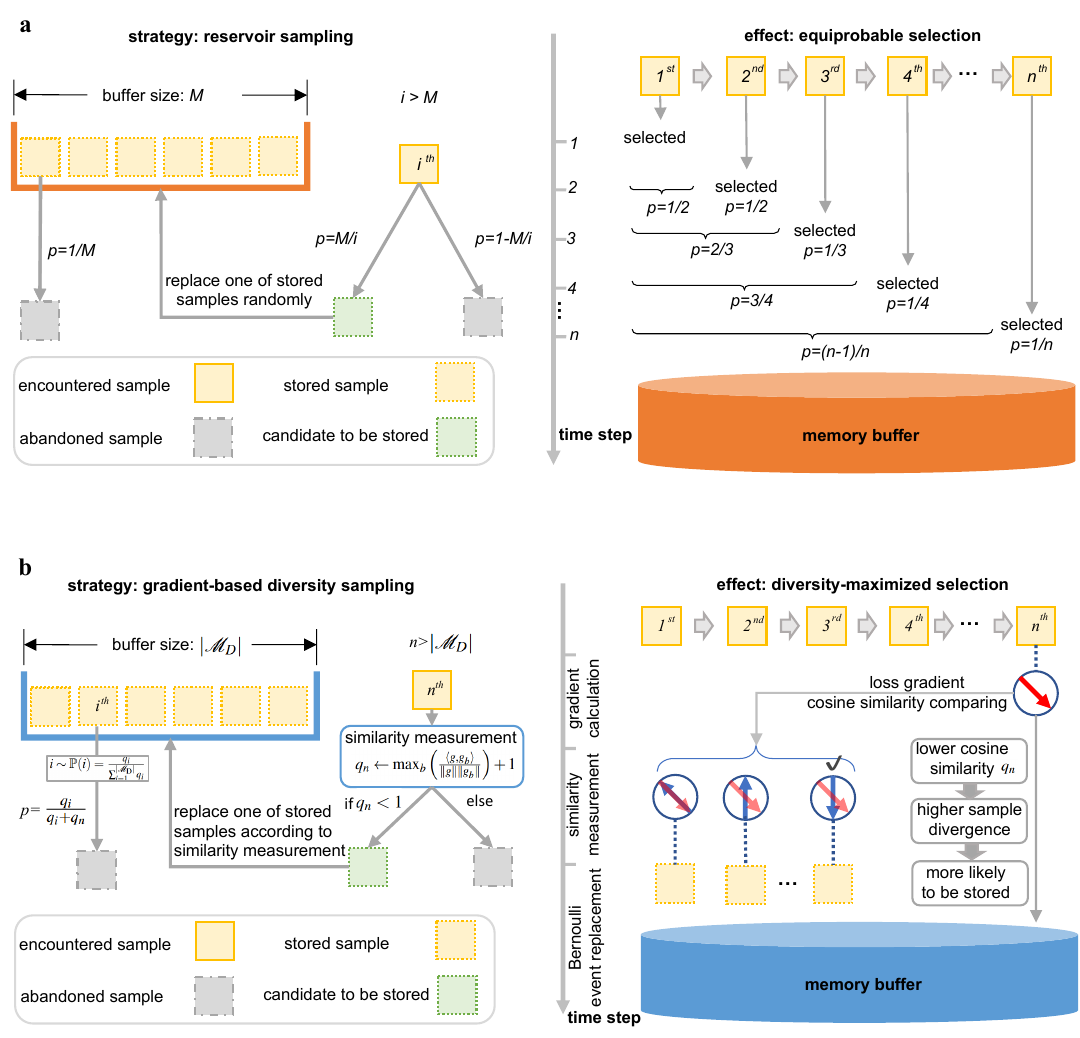}
\caption{\justifying Two memory sampling strategies in Dual-LS. \textbf{a}, Reservoir sampling strategy selects each sample from an unlimited online stream with equal probability: First, the buffer is initially empty. As new samples are observed, they are stored in the buffer until its capacity is reached. Once full, newly encountered samples may replace one of the existing samples. The probability of a sample being stored is designed to be uniform across all samples in the data stream. For each $n > |\mathcal{M}_\text{R}|$, the $n\textsuperscript{th}$ observed sample is selected as a candidate with probability $|\mathcal{M}_\text{R}|/n$. Once selected, one of the previously stored samples is randomly replaced with this candidate, with the replacement probability being $1/|\mathcal{M}_\text{R}|$. \textbf{b}, Gradient-based diversity sampling strategy aims to maximize the diversity of samples in the memory buffer: The diversity is quantified by the cosine similarity between the loss gradients of samples, referred to as the similarity score, $q$. The smaller $q$ for a sample represents the relatively higher diversity of this sample to others. Thus, in order to maximize the diversity of stored samples, a sample in the buffer with a larger $q$ value is more likely to be replaced by a newly encountered sample. Let the similarity score of the new sample be denoted as $q_n$, and the similarity score of the candidate sample be $q_i$. The replacement follows a Bernoulli distribution, with the probability of replacement given by $q_i / (q_n + q_i)$.}
\label{fig6_two_buffer}
\end{figure}

Fig.~\ref{fig5_CLS}a illustrates the distribution of samples from different past tasks in two memory replay buffers when the buffer size is set to 1,000. The reservoir sampling algorithm (Fig.~\ref{fig5_CLS}a bottom) ensures that the selection probability of all samples is unaffected by the sequence of data in the stream. Consequently, the proportion of samples from each past task in the memory buffer is relatively balanced and aligns with the proportion of training samples in each past scenario. The detailed process of reservoir sampling strategy for $\mathcal{M}_\text{R}$ is depicted in Algorithm~\ref{al_1} in \textbf{Extended Data}. Initially, the buffer is empty. As new samples are observed, they are stored in the buffer. Once the buffer is full, newly encountered samples may replace one of the existing samples. Notably, the probability of a sample being stored is intended to be uniform across all samples in the data stream. However, the total length of the data stream, $|\mathcal{D}|$, is unknown, as the samples arrive incrementally. To realize this equiprobable selection from the data stream, we applied reservoir sampling. As depicted in Fig.~\ref{fig6_two_buffer}a, for every $n > |\mathcal{M}_\text{R}|$, the $n\textsuperscript{th}$ observed sample is selected as a candidate with a probability of $|\mathcal{M}_\text{R}|/n$. Once the $n\textsuperscript{th}$ sample is chosen as the candidate, one of the previously stored samples is randomly replaced with this candidate, with the replacement probability being $1/|\mathcal{M}_\text{R}|$. Consequently, each sample in $\mathcal{D}$ has an equal probability of $|\mathcal{M}_\text{R}|/|\mathcal{D}|$ of being stored in the buffer.

In contrast, the diversity-aware memory buffer aims to enhance the diversity of samples stored in the buffer. As described in Algorithm~\ref{al_2} in \textbf{Extended Data}, the gradient-based diversity sampling aims at maximizing the diversity among the selected samples. The diversity is quantified by the cosine similarity between the loss gradients of samples\cite{aljundi2019gradient}, referred to as the similarity score $q$. A sample in the buffer with a larger value of $q$ is more likely to be replaced by a newly encountered sample. When the buffer is not yet full, each incoming sample $\mathbf{d}_n$ is stored along with its associated similarity score $q_n$. Once the buffer reaches its capacity, a sample is randomly selected from the buffer to serve as the candidate for replacement. Let the similarity score of the new sample be denoted as $q_n$, and the similarity score of the candidate sample be $q_i$. The replacement process follows a Bernoulli distribution, where the probability of replacement is given by $q_i / (q_n + q_i)$. Fig.~\ref{fig6_two_buffer}b also shows the mechanism of the gradient-based diversity sampling. As shown in the top of Fig.~\ref{fig5_CLS}a, as the model progresses from Task 1 to Task 8, the proportion of samples from each past task in the buffer does not strictly match the proportion of training samples in the dataset. Specifically, more training samples from Task 7 and Task 8 are retained in the buffer, whereas fewer samples from Tasks 1 through 6 are preserved by the end of the learning process. This comparison indicates that training samples from Task 7 and Task 8 exhibit higher diversity than those from Tasks 1 to 6. This characteristic facilitates the selection of more representative samples to construct the memory buffer. Fig.~\ref{fig5_CLS}b shows the sampling results for each past task when the buffer sampling size is set to 8, further validating the conclusions drawn from Fig.~\ref{fig5_CLS}a. The complementary strengths of the two memory buffers can be combined to enhance model performance. This mechanism resembles the interplay between the hippocampus and neocortex in the human brain, which improves learning efficiency and outcomes.

In addition to the memory buffer design being inspired by biological intelligence, the learning mechanism of Dual-LS also embodies the principles of the CLS theory. Fig.~\ref{fig5_CLS}c illustrates the interplay between short-term and long-term models. For samples retrieved from the memory buffer, comparing the fast model's error with the slow model's error enables effective selection of the current model's representation in prediction. The detailed mechanism of the interplay between the fast and slow models is provided in Methods. Fig.~\ref{fig5_CLS}d shows the training loss, indicating that during the training process under the CL paradigm, the model's predictive performance declines after task transitions.

In Fig.~\ref{fig5_CLS}e, we present a case study to demonstrate the impact of CL on the vehicle motion forecasting. The target vehicle is expected to continue moving along the inner lane of the roundabout and reach the location marked by the green triangle after 3 seconds. This sample is part of the test set from task 5. After the model completes training on all samples from the fifth task, the proposed Dual-LS algorithm achieves a prediction similar to that of the vanilla model. However, as the model continues learning up to Task 8, Dual-LS maintains its predicted target position close to the ground truth. In contrast, the vanilla model produces a completely incorrect prediction, estimating that the target vehicle will leave the roundabout at the next exit. Such a misjudgment could significantly impact the dynamic risk assessment of SV. More case studies are presented in Supplementary Material (Section 4).

In the proposed Dual-LS algorithm, the hyperparameters influence the model's performance.  the impact of the number of epochs, the update probability of the long-short term models, and the replay memory loss coefficient on prediction performance are detailed in Supplementary Material (Section 3). Experimental results indicate that increasing the number of epochs does not significantly enhance the CL performance of the DNN. Furthermore, no clear patterns emerge between other hyperparameter combinations and evaluation metrics. This suggests that for specific DNN, the selection of relevant CL hyperparameters requires tailored analysis.

% \begin{figure}[t!]
%     \centering
%     \includegraphics[width=1\linewidth]{figure/FT_case_2-2.pdf}
% \caption{\justifying Case 3 in the test set from scenario FT (task 2). Top: the vanilla model trained after 5-8 tasks, and tested in the 3\textsuperscript{rd} testing case in task 2. Down: the Dual-LS trained after 5-8 tasks, and tested in the 3\textsuperscript{rd} testing case in task 2. The TV with its observed trajectory is colored in red, and the background vehicles are colored in blue. The ground truth is represented by the green triangle, which is the position of the TV after 3 s. In each case, the predicions are represented by six yellow stars in red heatmaps, where the specific predicted postions are sampled based on the predicted heatmap.}
%     \label{fig5_case_study}
% \end{figure}

% \begin{figure}[t!]
% \centering
%     \includegraphics[width=0.9\linewidth]{figure/fig_6_ablation.pdf}
% \caption{\justifying Performance evaluation by ablation study. Each row donates one pair of hyper-parameters $p_{\text{S}}$ and $p_{\text{F}}$, which are the update frequency of slow and fast models. Each column represents each value of epoch. Three bars in each sub-figures represents different pairs of replay loss weight in Eq.\eqref{dual_memory_eq_4} with $\alpha=\alpha_1=\alpha_2$ and $\beta=\beta_1=\beta_2$.}
% \label{fig6_ablation_study}
% \end{figure}

\section*{Discussion}

For both humans and intelligent systems in smart cities, it is crucial to preserve past learning experiences while adapting to dynamic and ever-changing environments. However, the DNN is prone to forget the learned knowledge if CL is not considered. Inspired by the memory replay mechanism in the human brain, this article proposes the OTFCL paradigm to address the issue of catastrophic forgetting during the learning process from continuous data stream. Experimental results indicate that in the motion forescasting for intelligent vehicles, incorporating CL into the pretrain-and-deploy pipeline improves the prediction performance. This enhancement benefits prediction-based risk assessment, contributing to the safety of SVs and other vulnerable road users\cite{xing2024comprehensive}. Furthermore, it promotes the harmony in human-machine interactions\cite{wang2022social}. The proposed paradigm is generalizable and applicable not only to SVs but also to data-driven industrial robots\cite{dai2024autonomous}, service robots\cite{zhou2022human}, and other intelligent systems\cite{meng2025preserving}. Additionally, the OTFCL paradigm imposes strict constraints on the storage capacity of the memory buffer, effectively preventing unlimited data storage from the data stream. This facilitates the development of AI-based smart cities with controllable data storage costs, addressing energy efficiency and environmental sustainability challenges\cite{stanojevic2024high,yin2023accurate}.

Experiments demonstrate that the CLS of the human brain can inspire the design of CL strategies\cite{jiao2024brain}. The integration of dual memory systems can mitigate the issue of sample imbalance across different past tasks in the OTFCL paradigm while ensuring that previously encountered samples are replayed with equal probability. Moreover, the coordination of long-term and short-term learning strategies enables a balance between long-term memory consolidation and the rapid acquisition of short-term knowledge. This design emerges the interplay mechanism between the hippocampus and the neocortex in the human brain. Furthermore, the partitioned architecture of the human brain may offer valuable insights for designing DNN-based CL frameworks.

AI and biological intelligence share common goals in the CL of new knowledge\cite{kudithipudi2022biological,wu2022brain,van2020brain}, the rapid consolidation of experiences, and the efficient coordination of memory systems. Emerging theories in human brain science, neuroscience, and memory science will continue to inspire the architectural design of DNN\cite{williamson2024learning,gava2024organizing,zaki2024offline,shuai2010forgetting}. As described in this article, the theory of CLS between the hippocampus and the neocortex reflected in the Dual-LS algorithm, which effectively mitigates catastrophic forgetting. Subsequent work could further explore the integration of CL and collective learning to investigate how individual biological intelligence is transferred, shared, and expanded within society\cite{soltoggio2024collective,hassabis2017neuroscience}.

\section*{Methods}
\subsection*{Problem definition}
In CL, the DNN-based learner $f$ encounters $T$ tasks $\{\mathcal{T}_1,...,\mathcal{T}_T\}$ sequentially. Each task $\mathcal{T}_t$ ($1\le t\le T$) has its learning objective $\mathcal{O}_t$ and contains $N_t$ training samples $\{\mathbf{X}_i,\mathbf{Y}_i\}_{i=1}^{N_t}$, which are drawn from an i.i.d. distribution $\mathcal{D}_t$. The learning goal within each task can be expressed as:
\begin{equation}
    \mathbf{\theta}_{t}^{*} = \arg\min_{\mathbf{\theta}_{t}}\mathcal{L}_{t} \quad \text{with} \quad \mathcal{L}_{t} \triangleq\mathbb{E}_{(\mathbf{X},\mathbf{Y})\sim\mathcal{D}_{t}}\left( \ell\left(f_{\mathbf{\theta}_{t}}\left( \mathbf{X}\right), \mathbf{Y}\right)\right)
    \label{eq_single_task_1}
\end{equation}
where $\mathcal{L}_{t}$ is the loss function of task $\mathcal{T}_t$ and $\mathbb{E}$ is the expectation in the calculation of loss. $\ell$ is each sample's loss between prediction $f_{\mathbf{\theta}_{t}}\left( \mathbf{X}\right)$ and ground truth $\mathbf{Y}$. $(\mathbf{X},\mathbf{Y})$ is the sample following the distribution $\mathcal{D}_t$. $f_{\boldsymbol{\theta}_{t}}$ is the learning function of task $\mathcal{T}_t$ equipped with parameters $\boldsymbol{\theta}_{t}$. Usually, different tasks correspond to distinct distributions, 
\begin{equation}
\forall t_1\neq t_2,\quad\mathcal{D}_{t_1} \neq \mathcal{D}_{t_2}
\end{equation}
According to the availability of the task identifier in testing and whether the model needs to infer it, the CL can be categorized into three types\cite{van2022three}: task-incremental learning (TIL), domain-incremental learning (DIL), and class-incremental learning (Class-IL). In this article, the output representation is fixed for all tasks and the model has no access to the task identifier in testing, so the vehicle motion forecasting is a variant of DIL from image classification. Based on Eq.~\eqref{eq_single_task_1}, the sequential learning process from first task $\mathcal{T}_1$ to the current task $\mathcal{T}_{t_c}$ is formulated as:
\begin{equation}
    \mathbf{\theta}_{1:t_c}^{*} = \arg\min_{\mathbf{\theta}_{1:t_c}} \sum_{t=1}^{t_c}\mathcal{L}_{t}  \quad \text{with} \quad \mathcal{L}_{t} \triangleq\mathbb{E}_{(\mathbf{X},\mathbf{Y})\sim\mathcal{D}_{t}}\left( \ell\left(f_{\boldsymbol{\theta}}\left( \mathbf{X}\right), \mathbf{Y}\right)\right)
    \label{eq_CL_task_1}
\end{equation}
where $\mathbf{\theta}_{1:t_c}^{*}$ is the optimized weights of DNN until the current task $\mathcal{T}_{t_c}$ and ${t_c}\in \{1,...,T\}$ is the index of current task. In Eq.~\eqref{eq_CL_task_1}, obtaining the optimal weights requires access to the dataset corresponding to each task that conforms to distribution $\{\mathcal{D}_1,...,\mathcal{D}_{t_c-1}\}$. However, in CL, it is not possible to access all the original data from previously experienced tasks, meaning that not all samples following distribution $\{\mathcal{D}_1,...,\mathcal{D}_{t_c-1}\}$ can be obtained. Ideally, the optimization process in Eq.~\eqref{eq_CL_task_1} aims to mimic the prediction performance on past samples while fitting the samples in the current task $\mathcal{T}_{t_c}$. In this way, the model is designed to approximate the best and original prediction for past samples. Therefore, the learning process is converted into:
\begin{equation}\label{eq_CL_task_2}
   \mathbf{\theta}_{1:t_c}^{*} = \arg\min_{\mathbf{\theta}_{1:t_c}} \mathcal{L}_{t_c} + \alpha \sum_{t=1}^{t_c-1} \mathbb{E}_{(\mathbf{X},\mathbf{Y})\sim\mathcal{D}_{t}} \left[ D_{\text{KL}}\left( f_{{\mathbf{\theta}}^{*}_{1:t}}(\mathbf{X}) \, \| \, f_{{\mathbf{\theta}}_{1:t_c}}(\mathbf{X}) \right) \right]
\end{equation}
where $\alpha$ is the hyper-parameter to trade off the current task and all previous tasks. $D_{\text{KL}}$ is the Kullback–Leibler divergence between the best response of sample input $\mathbf{X}$ with ${\mathbf{\theta}}_{1:t_c}$ and optimal parameters ${\mathbf{\theta}}^{*}_{1:t}$. In the vehicle motion forecasting of SVs, the input feature combines dynamic information from traffic participants and static information from the road map:
\begin{equation}\label{eq_input}
    \mathbf{X}=\{ \mathbf{O}, \mathbf{E}\}
\end{equation}
where $\mathbf{O}\in \mathbb{R}^{d_\text{v} \times d_\text{s}}$ and $\mathbf{E} \in \mathbb{R}^{d_\text{e}}$ donate features for dynamic and static inputs, respectively. $d_\text{v}$ is the number of vehicles in the scenario and $d_\text{s}$ is the length of feature space for each vehicle. The output of DNN $\hat{\mathbf{Y}}\in \mathbb{R}^{l\times w}$ is represented by the fixed-size heatmap, which is the non-parametric distribution of the two-dimensional goal position. $l$ and $w$ are the length and width of the rectangular heatmap, respectively. The prediction of goal distribution can be formulated as:
\begin{equation}
    \hat{\mathbf{Y}}=f_{\mathbf{\theta}}(\mathbf{X})=f_{\mathbf{\theta}}(\mathbf{O},\mathbf{E})
\end{equation}
The structure of the DNN for vehicle motion forecasting is detailed in the Supplementary Material (Section 2).

\subsection*{Replay-based continual learning}
For most data-driven learning processes of DNN, the costs associated with data storage and maintenance are substantial, especially when effective they need access to all previously encountered samples. Furthermore, as the number of tasks increases, methods that indiscriminately revisit a large volume of past task samples will degrade the training efficiency. This is unacceptable for models that require online learning and rapid updates. In CL, a common strategy is to maintain only a small subset of past samples to represent historical tasks. This collection of limited past samples is referred to as the memory buffer $\mathcal{M}$. Thus, the Eq.~\eqref{eq_CL_task_2} can be transformed into:
\begin{equation}\label{eq_CL_task_3}
   \mathbf{\theta}_{1:t_c}^{*} = \arg\min_{\mathbf{\theta}_{1:t_c}} \mathcal{L}_{t_c} + \alpha \sum_{t=1}^{t_c-1} \mathbb{E}_{(\mathbf{X},\mathbf{Y})\sim\mathcal{M}_{t}} \left[ D_{\text{KL}} \left( f_{{\mathbf{\theta}}^{*}_{1:t}}(\mathbf{X}) \, \| \, f_{{\mathbf{\theta}}_{1:t_c}}(\mathbf{X}) \right) \right]
\end{equation}
where $\mathcal{M}_{t}$ is the memory buffer of $t^\textsuperscript{th}$ task. it indicates that the learner can select samples from $\mathcal{M}_{t}$. To guarantee the efficiency of CL, the size of $\mathcal{M}$ is strictly limited:
\begin{equation}
   |\mathcal{M}| \ll \sum_{t=1}^{T}N_t \label{eq_amount}
\end{equation}
where $|\cdot|$ is the size of memory buffer. The maintenance of the memory buffer and the sampling process from it are collectively termed replay-based CL. In Eq.~\eqref{eq_CL_task_3}, the ground truth label (goal position) is not considered in the approximation to past samples, which is important information in measuring the performance. Following the operation in the last term of Eq.~\eqref{eq_CL_task_3}, an improved expression is as follow\cite{buzzega2020dark}:
\begin{equation}\label{eq_CL_task_4}
\mathbf{\theta}_{1:t_c}^{*} = \arg\min_{\mathbf{\theta}_{1:t_c}} \mathcal{L}_{t_c} + \alpha \sum_{t=1}^{t_c-1} \mathbb{E}_{(\mathbf{X},\mathbf{Y})\sim\mathcal{M}_{t}} \left[ D_{\text{KL}}  \left( f_{{\mathbf{\theta}}^{*}_{1:t}}(\mathbf{X}) \, \| \, f_{{\mathbf{\theta}}_{1:t_c}}(\mathbf{X}) \right) \right] +\beta \sum_{t=1}^{t_c-1} \mathbb{E}_{(\mathbf{X},\mathbf{Y})\sim\mathcal{M}_{t}} \left[ \ell\left(f_{\mathbf{\theta}_{1:t_c}}\left( \mathbf{X}\right), \mathbf{Y}\right) \right]
\end{equation}
where $\beta$ is the hyper-parameter to trade off the influence of ground truth-based loss. 

\subsection*{Online task-free continual learning paradigm (OTFCL)}
Eq.~\eqref{eq_CL_task_4} represents the standard form of CL with task constraints. However, in the fields of vehicle motion forecasting, not all scenarios have task identifiers, which can be manually specified as strong constraints in CL. In many cases, obtaining task identifiers is impractical. According to Eq.~\eqref{eq_CL_task_4}, if the task guidance is absent, it can be expressed in a more general task-free form:
\begin{equation}\label{eq_CL_task_5}
\mathbf{\theta}_{\text{TF}}^{*} = \arg\min_{\mathbf{\theta}_{\text{TF}}} \mathcal{L}_{t_c} + \alpha  \mathbb{E}_{(\mathbf{X},\mathbf{Y})\sim\mathcal{M}} \left[ D_{\text{KL}}  \left( f_{{\mathbf{\theta}}^{*}}(\mathbf{X}) \, \| \, f_{{\mathbf{\theta}}_{\text{TF}}}(\mathbf{X}) \right) \right] +\beta\mathbb{E}_{(\mathbf{X},\mathbf{Y})\sim\mathcal{M}} \left[ \ell\left(f_{\mathbf{\theta}_{\text{TF}}}\left( \mathbf{X}\right), \mathbf{Y}\right) \right]
\end{equation}
where $\mathbf{\theta}_{\text{TF}}$ are the weights of the model with the task-free CL setting. The task-based memory buffer $\mathcal{M}_{t}$ is relaxed as $\mathcal{M}$, which stores selected samples from the stream data. It is the form of general continual learning (GCL)\cite{buzzega2020dark} with the following characteristics: no task label, no test time oracle, and limited memory buffer size. 

To realize OTFCL based on GCL, four challenges need to be addressed: First, for a learner with online learning and updating capabilities, the samples selected from the memory buffer and those from the data stream can only be used for one backpropagation iteration, preventing multiple epochs in the training process. Second, from the perspective of efficient training in CL, the model cannot save the best model in real-time after training on samples obtained from the data stream. Therefore, the key information corresponding to the samples needs to be retained during the training process. Third, it is essential to select effective and representative samples from the stream data to build the memory buffer. Fourth, after constructing the memory buffer, an effective learning strategy is required under the setting of OTFCL. The first challenge is the basic setting in OTFCL, which puts a higher demand on one-shot learning. The second one is solved by the heatmap-based output representation. The last two issues are considered in the proposed Dual-LS algorithm.

\subsection*{Complementary Learning System theory in continual learning}
The core of CLS theory is that the human brain has two complementary subsystems for learning and remembering knowledge: the hippocampus is responsible for short-term rapid memory, while the neocortex works for long-term structured memory. The interplay between the hippocampus and the neocortex effectively enhances the learning process. Inspired by this biological mechanism, we developed a new algorithm, Dual-LS, based on replay-based CL methods, which integrates CLS theory into the design of the memory buffer and model update strategy. In the following sections, we will introduce the design of the dual memory replay buffer and the strategy for the long-short term.

\subsubsection*{Dual memory replay buffer}\label{dual_memory}

In the replay-based task-free CL\cite{wang2024comprehensive}, the most commonly used memory buffer selection method is reservoir sampling, which ensures that all samples from a dynamically distributed data stream have an equal probability of being selected. The detailed algorithm is shown in Extended Data Algorithm~\ref{al_1}, and the illustration of equal probability is provided in Fig.~\ref{fig6_two_buffer}a. The memory buffer obtained from reservoir sampling is denoted as $\mathcal{M}_{\text{R}}$. However, in the absence of task labels, if there is a data imbalance, the constraint of equal probability may lead to more frequent sampling from certain distributions. In other words, it will result in an imbalanced sample selection in the memory buffer. To trade off the samples across different distributions, the diversity is introduced into the memory buffer selection process. The goal is to address the imbalance in the data stream while ensuring sample diversity during buffer construction. Generally, for a data stream with $N$ samples, to simplify the problem, we assume only one sample can be observed at a time. When receiving the $n\textsuperscript{th}$ sample $n\in \{1, 2,..., N\}$,  the following objective needs to be achieved to maintain the performance of the DNN:
\begin{equation}\label{dual_memory_eq_1}
\begin{aligned}
\mathbf{\theta}_{\text{TF}}^{n} &= \arg\min_{\mathbf{\theta}_{\text{TF}}^{n}} \ell\left(f_{\mathbf{\theta}_{\text{TF}}^{n}}\left( \mathbf{X}_{n}\right), \mathbf{Y}_{n}\right)\\
&\text{s.t.}\quad \ell\left(f_{\mathbf{\theta}_{\text{TF}}^{n}}\left( \mathbf{X}_{i}\right), \mathbf{Y}_{i}\right)\le \ell\left(f_{\mathbf{\theta}_{\text{TF}}^{n-1}}\left( \mathbf{X}_{i}\right), \mathbf{Y}_{i}\right) \ \ \ \forall i\in\{1, 2,..., n-1\}
\end{aligned}
\end{equation}
where $f_{\mathbf{\theta}_{\text{TF}}^{n}}$ is the DNN model for OTFCL parametrized by $\mathbf{\theta}_{\text{TF}}^{n}$. $n$ and $i$ are the indexes of current and past samples from the data stream, respectively. As suggested in GEM, the constraints in Eq.~\eqref{dual_memory_eq_1} are equivalent to the constraints of gradients:
\begin{equation}\label{dual_memory_eq_2}
\left\langle \mathbf{g}_{n}, \mathbf{g}_{i}\right\rangle=\left\langle\frac{\partial \ell\left(f_{\mathbf{\theta}_{\text{TF}}^{n}}\left(\mathbf{X}_{n}\right), \mathbf{Y}_{n}\right)}{\partial {\mathbf{\theta}_{\text{TF}}^{n}}}, \frac{\partial \ell\left(f_{\mathbf{\theta}_{\text{TF}}^{n}}\left(\mathbf{X}_{i}\right), \mathbf{Y}_{i}\right)}{\partial {\mathbf{\theta}_{\text{TF}}^{n}}}\right\rangle \geq 0 \ \ \ \forall i\in\{1, 2,..., n-1\}
\end{equation}
where $\mathbf{g}_{n}$ and $\mathbf{g}_{i}$ is gradients for current and past samples. $\left\langle\cdot,\cdot\right\rangle$ is the high-dimensional angle between two vectors. To realize the objective in Eq.~\eqref{dual_memory_eq_1} by satisfying constraints in Eq.~\eqref{dual_memory_eq_2}, the optimization process can be converted into maximizing the variance of gradient directions:
\begin{equation}\label{dual_memory_eq_3}
\begin{aligned}
\mathcal{M}_{\text{D}}^{*}&=\arg\max_{\mathcal{M}_{\text{D}}}\operatorname{Var}_{\mathcal{M}_{\text{D}}}\left[\frac{\mathbf{g}}{\|\mathbf{g}\|}\right] \\
& =\arg\max_{\mathcal{M}_{\text{D}}}\left( \frac{1}{|\mathcal{M}_{\text{D}}|} \sum_{m \in \mathcal{M}_{\text{D}}}\left\|\frac{\mathbf{g}_{m}}{\|\mathbf{g}_{m}\|}\right\|^{2}-\left\|\frac{1}{|\mathcal{M}_{\text{D}}|} \sum_{m \in \mathcal{M}_{\text{D}}} \frac{\mathbf{g}_{m}}{\|\mathbf{g}_{m}\|}\right\|^{2} \right)\\
& =\arg\max_{\mathcal{M}_{\text{D}}} \left(1-\frac{1}{|\mathcal{M}_{\text{D}}|^{2}} \sum_{i, j\in \mathcal{M}_{\text{D}}} \frac{\left\langle \mathbf{g}_{i}, \mathbf{g}_{j}\right\rangle}{\left\|\mathbf{g}_{i}\right\|\left\|\mathbf{g}_{j}\right\|} \right)
\end{aligned}
\end{equation}
where $\mathcal{M}_{\text{D}}$ is the sampling diviersity-based memory buffer and   $|\mathcal{M}_{\text{D}}|$ is the number of samples in $\mathcal{M}_{\text{D}}$. $\|\cdot\|$ is the 2-norm for gradients. $m$, $i$ and $j$ are indexes of samples in $\mathcal{M}_{\text{D}}$. To improve the efficiency in maintaining the memory buffer$\mathcal{M}_{\text{D}}$, a greedy-based algorithm is applied, which is presented in Extended Data Algorithm~\ref{al_2}. The illustration of  gradient-based diversity sampling strategy are detailed in Fig.~\ref{fig6_two_buffer}b. With samples from two kinds of memory $\mathcal{M}_{\text{R}}$ and $\mathcal{M}_{\text{D}}$ , the learning process in Eq.~\eqref{eq_CL_task_5} is further updated as:
\begin{equation}\label{dual_memory_eq_4}
\begin{aligned}
\mathbf{\theta}_{\text{TF}}^{*} = \arg\min_{\mathbf{\theta}_{\text{TF}}^{n}} \mathcal{L}_{t_c} + &\alpha_{1}\mathbb{E}_{(\mathbf{X},\mathbf{Y})\sim\mathcal{M}_{\text{R}}} \left[ D_{\text{KL}}  \left( f_{{\mathbf{\theta}}^{*}}(\mathbf{X}) \, \| \, f_{{\mathbf{\theta}}_{\text{TF}}}(\mathbf{X}) \right) \right] +\beta_{1}\mathbb{E}_{(\mathbf{X},\mathbf{Y})\sim\mathcal{M}_{\text{R}}} \left[ \ell\left(f_{\mathbf{\theta}_{\text{TF}}}\left( \mathbf{X}\right), \mathbf{Y}\right) \right]\\
&\alpha_{2}\mathbb{E}_{(\mathbf{X},\mathbf{Y})\sim\mathcal{M}_{\text{D}}} \left[ D_{\text{KL}}  \left( f_{{\mathbf{\theta}}^{*}}(\mathbf{X}) \, \| \, f_{{\mathbf{\theta}}_{\text{TF}}}(\mathbf{X}) \right) \right] +\beta_{2}\mathbb{E}_{(\mathbf{X},\mathbf{Y})\sim\mathcal{M}_{\text{D}}} \left[ \ell\left(f_{\mathbf{\theta}_{\text{TF}}}\left( \mathbf{X}\right), \mathbf{Y}\right) \right]
\end{aligned}
\end{equation}
where $\alpha_{1}$, $\alpha_{2}$, $\beta_{1}$ and $\beta_{2}$ are parameters trading off different replay loss.

\subsubsection*{Long-short term model update strategy}\label{ls_update}
In addition to incorporating CLS theory into the design of the dual memory buffer, a fast-slow integrated model update mechanism is designed in Dual-LS to reflect the interplay between rapid and consolidated memory in the human brain when acquiring new knowledge. Specifically, the model $f_{\mathbf{\theta}_{\text{TF}}^{*}}$ trained directly using the data stream under the OTFCL paradigm is the working model $f_{\mathbf{\theta}_{\text{TF}}^{*}}^{\text{W}}$. Furthermore, based on the working model, we simultaneously maintain a fast learning model $f_{\mathbf{\theta}_{\text{TF}}^{*}}^{\text{F}}$ and a slow learning model $f_{\mathbf{\theta}_{\text{TF}}^{*}}^{\text{S}}$, corresponding to rapid and consolidated knowledge, respectively. The update of the working model is based on the Eq.~\eqref{dual_memory_eq_4}. Meanwhile, $f_{\mathbf{\theta}_{\text{TF}}^{*}}^{\text{F}}$ and $f_{\mathbf{\theta}_{\text{TF}}^{*}}^{\text{S}}$ are stochastically updated with fixed manually set probabilities $p_{\text{F}}$ and $p_{\text{S}}$. We make $p_{\text{F}}\textless p_{\text{S}}$ to guarantee that the fast model changes frequently. The specific strategy is taking an exponential moving average of the working model's weights:
\begin{equation}\label{eq_ls_update_1}
\mathbf{\theta}_{\text{TF}}^{\text{F}} = \alpha_{\text{F}}\mathbf{\theta}_{\text{TF}}^{\text{F}}+(1-\alpha_{\text{F}})\mathbf{\theta}_{\text{TF}}^{\text{W}}
\end{equation}
\begin{equation}\label{eq_ls_update_2}
\mathbf{\theta}_{\text{TF}}^{\text{S}} = \alpha_{\text{S}}\mathbf{\theta}_{\text{TF}}^{\text{S}}+(1-\alpha_{\text{S}})\mathbf{\theta}_{\text{TF}}^{\text{W}}
\end{equation}
where $\alpha_{\text{F}}$ and $\alpha_{\text{S}}$ are decaying rates. Additionally, to reflect the enhancement effect of the fast model and slow model on the working model, the memory buffer samples used to update the working model will select the heatmap with the smallest prediction error for the calculation of KL divergence in Eq.~\eqref{dual_memory_eq_4}:
\begin{equation}\label{eq_ls_select_sample}
f_{\mathbf{\theta}^{*}}\left( \mathbf{X}_{\text{M}}\right) = \left\{\begin{array}{l}
f_{\mathbf{\theta}^{*}}^{\text{F}}\left( \mathbf{X}_{\text{M}}\right)\quad\textbf{if} \quad \ell\left(f_{\mathbf{\theta}^{*}}^{\text{F}}\left( \mathbf{X}_{\text{M}}\right), \mathbf{Y}_{\text{M}}\right) \textless \ell\left(f_{\mathbf{\theta}^{*}}^{\text{S}}\left( \mathbf{X}_{\text{M}}\right), \mathbf{Y}_{\text{M}}\right)\\
f_{\mathbf{\theta}^{*}}^{\text{S}}\left( \mathbf{X}_{\text{M}}\right) \quad \textbf{other}
\end{array} \right.
\end{equation}
where $\mathbf{X}_{\text{M}}$ and $\mathbf{Y}_{\text{M}}$ are samples selected from the memory buffer. The complete diagram of the long short term model update is detailed in Extended Data Algorithm~\ref{al_3}.

\subsection*{Implementation}
In this paper, we mainly consider the DIL setting in OTFCL paradigm. To verify the proposed Dual-LS algorithm, the interaction dataset is selected, which covers intersections, roundabouts, and highway merging. In experiments, each traffic scenario corresponds to a task. The duration of the original recorded video, the number of samples, and the corresponding order in data streaming are detailed in the Supplementary Material (Section 1). The proposed OTFCL paradigm and dual-LS algorithm both require a basic DNN model in specific prediction tasks. Our previous work uncertainty quantification network (UQnet) is selected as the base network for vehicle motion forecasting\cite{li2024UQnet}, which is the 1\textsuperscript{st} SOTA algorithm for the interaction prediction challenge in ICCV 2021\footnote{INTERPRET: Interaction-dataset-based prediction challenge, single-agent track, organized by ICCV 2021. Available at: \url{https://challenge.interaction-dataset.com/leader-board/}. Last accessed on May 3\textsuperscript{rd}, 2025.}. As shown in Eq.~\eqref{eq_input}, the input of UQnet includes the historical trajectory information of surrounding vehicles in 1 second and the static map information, and the output is the non-parametric probability distribution of the target position after 3 seconds, which sums to 1. The details of the UQnet are introduced in the Supplementary Material (Section 2). In the training process, UQnet uses focal loss to quantify the differences between non-parametric distribution and sparse target positions. 

\subsection*{Baseline algorithms}
To verify the effectiveness of the proposed Dual-LS algorithm, the following algorithms are selected as baselines.
\begin{itemize}
\item Vanilla\cite{li2024UQnet}: Vanilla  means that the UQnet trained sequentially from 1$\textsuperscript{st}$ to 8$\textsuperscript{th}$ scenarios. No CL strategy is applied to mitigate the catastrophic forgetting in DNN.
\item GEM\cite{lopez2017gradient}: GEM is a typical replay-based CL algorithm, which needs the guidance of task label. GEM equally allocates memory resource to each past task and the gradient update directions from each task's replay buffer are integrated together with an inequality constraint.
\item A-GEM\cite{chaudhry2018efficient}: A-GEM is the modified version of GEM, which relaxes the setting of task-based to task-free.
\item DER\cite{buzzega2020dark}: A replay-based general CL method, which uses reservoir sampling to maintain the memory buffer.
\item GSS\cite{aljundi2019gradient}: A replay-based task-free CL method, which applies the gradient-based sample selection in the memory buffer.
\end{itemize}

\subsection*{Evaluation metrics}
To evaluate the prediction accuracy of the proposed approach, this work adopts two metrics\cite{li2024UQnet}: the minimum final displacement error (FDE) and the miss rate (MR). The UQnet generates $K$ predicted goals, where the $k^\textsuperscript{th}$ position of $i^\textsuperscript{th}$ sample is denoted as $\hat {\mathbf{p}}_k^i$. FDE measures the minimum Euclidean distance between the $K$ predicted positions and the ground truth $\mathbf{p}$ over each sample. For the $i^\textsuperscript{th}$ sample, FDE is formulated as:
\begin{equation}
    \text{FDE}^i_K =\min_{k\in\left \{ 1,...,K \right \} }{\left\| \hat {\mathbf{p}}_{k}^{i} - \mathbf{p}^{i} \right\|}_2 \label{eq_fde_sample}
\end{equation}
where $\left\|\cdot \right\|_2$ is the measurement of Euclidean distance. For each task $\mathcal{T}_t$ ($1\le t\le T$), the  FDE is mean of all samples in the test set,
\begin{equation}
    \text{FDE}_{t}=\frac{1}{N_{t}^{\prime}}\sum_{i=1}^{N_{t}^{\prime}}\text{FDE}^i_K \label{eq_fde_task}
\end{equation}
where $N_{t}^{\prime}$ is the number of samples in the test set of task $\mathcal{T}_t$. The average FDE over all tasks is as follow:
\begin{equation}
    \text{FDE}_{\text{Ave}}=\frac{1}{T}\sum_{t=1}^{T}\text{FDE}_{t}
\end{equation}

The MR measures the percentage of predicted goals that are out of the rectangular area around the ground truth. The lateral threshold of the given area is 1 m, and the longitudinal threshold is determined by the piece-wise function depending on the velocity $v$ of the predicted vehicle:
\begin{equation}
th_{\text{MR}}(v) = \left\{ \begin{array}{l}
1, v < 1.4 \\
1 + \frac{{v - 1.4}}{{11 - 1.4}}, 1.4 \le v \le 11\\
2, v > 11
\end{array} \right.
\end{equation}
where the unit of the longitudinal threshold is $\rm{m}$, and the unit of the velocity $v$ is $\rm{ms^{-1}}$. Let $N^{\text{miss}}_{t}$ represent the total number of predicted goals that are out of the given area, the MR in the testing set of task $\mathcal{T}_t$ is $\frac{N^{\text{miss}}_{t}}{N_{t}^{'} \times W}$. The average MR over $T$ tasks can be represented as:

\begin{equation}
    \text{MR}_{\text{Ave}} = \frac{1}{T} \sum_{t=1}^{T} \frac{N^{\text{miss}}_{t}}{N_{t}^{\prime} \times W}
\end{equation}

Since FDE and MR both measure the prediction errors, the smaller values of FDE and MR represent the better prediction performance.

For the evaluation of the capability of CL, FDE-based backward transfer (FDE-BWT) and MR-based backward transfer (MR-BWT) are proposed to quantify catastrophic forgetting in CL tasks. If the prediction errors at the $j^\textsuperscript{th}$ task after training on the $i^\textsuperscript{th}$ task is denoted as $R_{i,j}$, FDE-BWT and MR-BWT can be computed uniformly by:
\begin{equation}
\text{FDE-BWT}=\frac{1}{t-1} \sum_{i=1}^{t-1}\left(\text{FDE}_{c,i}-\text{FDE}_{i,i}  \right) \quad \forall t \in \{1,...,T\}  
\end{equation}
\begin{equation}
\text{MR-BWT}=\frac{1}{t-1} \sum_{i=1}^{t-1}\left(\text{MR}_{c,i}-\text{MR}_{i,i}  \right) \quad \forall t \in \{1,...,T\}  
\end{equation}

Similar to FDE and MR, the smaller FDE-BWT and MR-BWT represent better CL performance to avoid catastrophic forgetting.

\bibliography{ref.bib}

\section*{Acknowledgements}
This research is supported by the National Natural Science Foundation of China under Grant 52372405.

\section*{Data availability}
The dataset used in this study is publicly available on \hyperlink{https://interaction-dataset.com/}{https://interaction-dataset.com/}.
\section*{Code availability}
The code used in this study is publicly available on \hyperlink{https://github.com/lzrbit/Dual-LS}{https://github.com/lzrbit/Dual-LS}.

\section*{Author contributions statement}
Z.L., C.L., and J.G. led the research project, and wrote the paper. Z.L., Y.L., and C.G developed the framework and algorithm. G.D., X.Z., C.L., and C.G. wrote the paper. All authors provided feedback during the manuscript writing and results discussions. C.L., and J.G. approved the submission and accepted responsibility for the overall integrity of the paper.

\section*{Additional information}
\textbf{Supplementary information} The online version contains supplementary material available at the website.

\noindent \textbf{Correspondence} and requests for materials should be addressed to Chao Lu and Jianwei Gong.

\clearpage
\section*{Extended data}\label{chap:extended data}
\SetAlgorithmName{Extended Data Algorithm}{Extended Data algorithms}{Extended Data algorithm}
\begin{algorithm}[h!]
\SetAlgoNlRelativeSize{0} % 保持行号大小不变
%\SetAlgoNlRelativeSize{-1} % 调整行号大小
\SetCommentSty{textnormal} % 注释字体与正文一致
\SetKwComment{tcp}{$\triangleright$}{}
\caption{Reservoir sampling strategy}\label{al_1}
\KwIn{The data stream $\mathcal{D}$, where the $n\textsuperscript{th}$ sample is denoted as $\mathbf{d}_n$; the rapid memory buffer $\mathcal{M}_\text{R}$ with maximum size $|\mathcal{M}_\text{R}|\in\mathbb{Z}^+$.}
\KwOut{The updated $\mathcal{M}_\text{R}$, where the $i\textsuperscript{th}$ sample stored in $\mathcal{M}_\text{R}$ is denoted as $\mathcal{M}_\text{R} (i)$.}

$\mathcal{M}_\text{R} \gets \emptyset$\tcp*[r]{Initialization.}

\For{$n$ \textbf{in} range(1, $|\mathcal{D}|$)}{
    \eIf{$n \leq |\mathcal{M}_\text{R}|$}{
        $\mathcal{M}_\text{R}(n) \gets \mathbf{d}_n$\,  
 $\mathcal{M}_\text{R} \gets \mathcal{M}_\text{R} \cup \mathcal{M}_\text{R}(n)$
    }{
        $r \sim \text{Uniform}(1, n)$\;
        \If{$r \leq |\mathcal{M}_\text{R}|$}{
            $\mathcal{M}_\text{R}(r) \gets \mathbf{d}_n$\tcp*[r]{Sample replacement.}
        }
    }
}
\end{algorithm}

\begin{algorithm}[h!]
\SetAlgoNlRelativeSize{0} % 保持行号大小不变
%\SetAlgoNlRelativeSize{-1} % 调整行号大小
\SetCommentSty{textnormal} % 注释字体与正文一致
\SetKwComment{tcp}{$\triangleright$}{}
\caption{Gradient-based diversity sampling strategy}\label{al_2}
\KwIn{The data stream $\mathcal{D}$, where the $n\textsuperscript{th}$ sample is denoted as $\mathbf{d}_n$ with the similarity score $q_n$; the distinctive memory buffer $\mathcal{M}_\text{D}$ with the maximum size $|\mathcal{M}_\text{D}|$; the number of samples $B$ for score computation; the loss function $\ell$; the vehicle motion forecasting model ${f_{\boldsymbol{\theta}_\text{TF}}}$.}
\KwOut{The updated $\mathcal{M}_\text{D}$, where the $i\textsuperscript{th}$ stored sample is denoted as $\mathcal{M}_\text{D}(i)$.}

\For{$n$ \textbf{in} range(1, $|\mathcal{D}|$)}{
    \eIf{$n == 1$}{
        $q_n \gets 0.1$\tcp*[r]{Initialization.}
    }{
        $\mathbf{g} \gets \nabla \ell(f_{\boldsymbol{\theta}_\text{TF}}(\mathbf{d}_n))$, $\Omega \gets \emptyset$\;
        
        \For{$k$ \textbf{in} range(1, $B$)}{
            $b \sim \text{Uniform}(1, |\mathcal{M}_\text{D}|)$,
            $\Omega \gets \Omega \cup \mathcal{M}_\text{D}(b)$\tcp*[r]{Randomly select $B$ stored samples for the score computation.}
        }
        
        \For{all samples in $\Omega$}{
            $\mathbf{g}_b \gets \nabla \ell(f_{\boldsymbol{\theta}_\text{TF}}(\mathcal{M}_\text{D}(b)))$\;
        }
        
        $q_n \gets \max_b \left( \frac{\langle \mathbf{g}, \boldsymbol{g}_b \rangle}{\|\mathbf{g}\|\|\boldsymbol{g}_b\|} \right) + 1$\tcp*[r]{Similarity score.}
        
        \eIf{$n \leq |\mathcal{M}_\text{D}|$}{
            $\mathcal{M}_\text{D} \gets \mathcal{M}_\text{D} \cup \mathbf{d}_n$\;
        }
        { \If{$q_n < 1$}{
            $i \sim \mathbb{P}(i) = \frac{q_i}{\sum_{i=1}^{|\mathcal{M}_\text{D}|} q_i}$,
            $r \sim \text{Uniform}(0, 1)$\;
            \If{$r < \frac{q_i}{q_i + q_n}$}{
                $\mathcal{M}_\text{D}(i) \gets \mathbf{d}_n$,
                $q_i \gets q_n$\tcp*[r]{Sample replacement.}}
            }
        }
    }
}
\end{algorithm}

\begin{algorithm}[t!]
\SetAlgoNlRelativeSize{0} % 保持行号大小不变
%\SetAlgoNlRelativeSize{-1} % 调整行号大小
\SetCommentSty{textnormal} % 注释字体与正文一致
\SetKwComment{tcp}{$\triangleright$}{}
\caption{Long-short term model update strategy}\label{al_3}
\KwIn{The data stream $\mathcal{D}$, memory buffer $\mathcal{M}_\text{R}$ and $\mathcal{M}_\text{D}$ Learning rate $\alpha_{\text{W}}$, decaying rates $\alpha_{\text{F}}$ and $\alpha_{\text{S}}$, Update rates $p_{\text{F}}$ and $p_{\text{S}}$}
\KwOut{$\mathbf{\theta}_{\text{TF}}^{\text{F}}$, $\mathbf{\theta}_{\text{TF}}^{\text{S}}$ and $\mathbf{\theta}_{\text{TF}}^{\text{W}}$.}

$\mathbf{\theta}_{\text{TF}}^{\text{F}}=\mathbf{\theta}_{\text{TF}}^{\text{S}}=\mathbf{\theta}_{\text{TF}}^{\text{W}}$\tcp*[r]{Initialization.}
\While{Training}{
$(\mathbf{X},\mathbf{Y})\sim\mathcal{D}$\;
$(\mathbf{X}_{\text{M}},\mathbf{Y}_{\text{M}})\sim\mathcal{M}_\text{R},\mathcal{M}_\text{D}$\;
$\hat{\mathbf{Y}}^{\text{F}}_{\text{M}}=f_{\mathbf{\theta}^{*}}^{\text{F}}(\mathbf{X}_{\text{M}})$, $\hat{\mathbf{Y}}^{\text{S}}_{\text{M}}=f_{\mathbf{\theta}^{*}}^{\text{S}}(\mathbf{X}_{\text{M}})$ \;
$\hat{\mathbf{Y}}_{\text{M}}\leftarrow \hat{\mathbf{Y}}^{\text{F}}_{\text{M}}\quad\textbf{if} \quad \ell\left(f_{\mathbf{\theta}^{*}}^{\text{F}}\left( \mathbf{X}_{\text{M}}\right), \mathbf{Y}_{\text{M}}\right) \textless \ell\left(f_{\mathbf{\theta}^{*}}^{\text{S}}\left( \mathbf{X}_{\text{M}}\right), \mathbf{Y}_{\text{M}}\right) \quad \textbf{else} \quad \hat{\mathbf{Y}}^{\text{S}}_{\text{M}}$ as in \eqref{eq_ls_select_sample}\;
Calculate total $\mathcal{L}$ as in \eqref{dual_memory_eq_4}\;
$\mathbf{\theta}_{\text{TF}}^{\text{W}}\leftarrow\mathbf{\theta}_{\text{TF}}^{\text{W}}-\alpha_{\text{W}}\nabla_{\mathbf{\theta}_{\text{TF}}^{\text{W}}}\mathcal{L} $\;
$a,b \sim \text{Uniform}(0, 1)$\;
$\mathbf{\theta}_{\text{TF}}^{\text{F}} = \alpha_{\text{F}}\mathbf{\theta}_{\text{TF}}^{\text{F}}+(1-\alpha_{\text{F}})\mathbf{\theta}_{\text{TF}}^{\text{W}}\quad \textbf{if}\quad a\textless p_{\text{F}}\quad\textbf{else}\quad\mathbf{\theta}_{\text{TF}}^{\text{F}}$ \;
$\mathbf{\theta}_{\text{TF}}^{\text{S}} = \alpha_{\text{S}}\mathbf{\theta}_{\text{TF}}^{\text{S}}+(1-\alpha_{\text{S}})\mathbf{\theta}_{\text{TF}}^{\text{W}}\quad \textbf{if}\quad b\textless p_{\text{S}}\quad\textbf{else}\quad\mathbf{\theta}_{\text{TF}}^{\text{S}}$ \;
Update $\mathcal{M}_\text{R}$ as in \eqref{eq_ls_update_1}\;
Update $\mathcal{M}_\text{D}$ as in \eqref{eq_ls_update_2}\;
}
\end{algorithm}

\end{document}